\pdfoutput=1

\documentclass[11pt]{article}

\usepackage[]{acl}

\usepackage{times}
\usepackage{latexsym}

\usepackage{multirow}
\usepackage{tabularx}
\usepackage{graphicx}
\usepackage{caption}
\usepackage{subcaption}
\usepackage{lipsum}  
\usepackage{pdfpages}
\usepackage{amsmath}
\usepackage{amssymb}
\usepackage{dblfloatfix}
\usepackage{CJK}
\usepackage{cleveref}
\usepackage{lipsum} 

\usepackage[T1]{fontenc}

\usepackage[utf8]{inputenc}

\usepackage{microtype}

\definecolor{mypink}{rgb}{0.858, 0.188, 0.478}

\definecolor{mygreen}{rgb}{0.13, 0.54, 0.13}

\title{Prompter: Zero-shot Adaptive Prefixes for Dialogue State Tracking Domain Adaptation}

\author{Taha Aksu\textsuperscript{$\dagger\ddag * $,}, Min-Yen Kan\textsuperscript{$\dagger$}, Nancy F. Chen\textsuperscript{$\ddag$} \\
 \textit{$\dagger$} National University of Singapore (NUS), Singapore\\
 \textit{$\ddag$} Institute for Infocomm Research ($I^2R$), A*STAR, Singapore \\
 \texttt{*taksu@u.nus.edu}}
\begin{document}
\maketitle
\begin{abstract}
 A  challenge in the Dialogue State Tracking (DST) field is adapting models to new domains without using any supervised data --- zero-shot domain adaptation. Parameter-Efficient Transfer Learning (PETL) has the potential to address this problem due to its robustness. However, it has yet to be applied to the zero-shot scenarios, as it is not clear how to apply it unsupervisedly. 

Our method, Prompter, uses descriptions of target domain slots to generate dynamic prefixes that are concatenated to the key and values at each layer's self-attention mechanism. This allows for the use of prefix-tuning in zero-shot. Prompter outperforms previous methods on both the MultiWOZ and SGD benchmarks. In generating prefixes, our analyses find that Prompter not only utilizes the semantics of slot descriptions but also how often the slots appear together in conversation. Moreover, Prompter's gains are due to its improved ability to distinguish ``none''-valued dialogue slots, compared against baselines.
\end{abstract}

\section{Introduction}

\begin{figure}[!ht]
\centering

\includegraphics[width=0.4\textwidth]{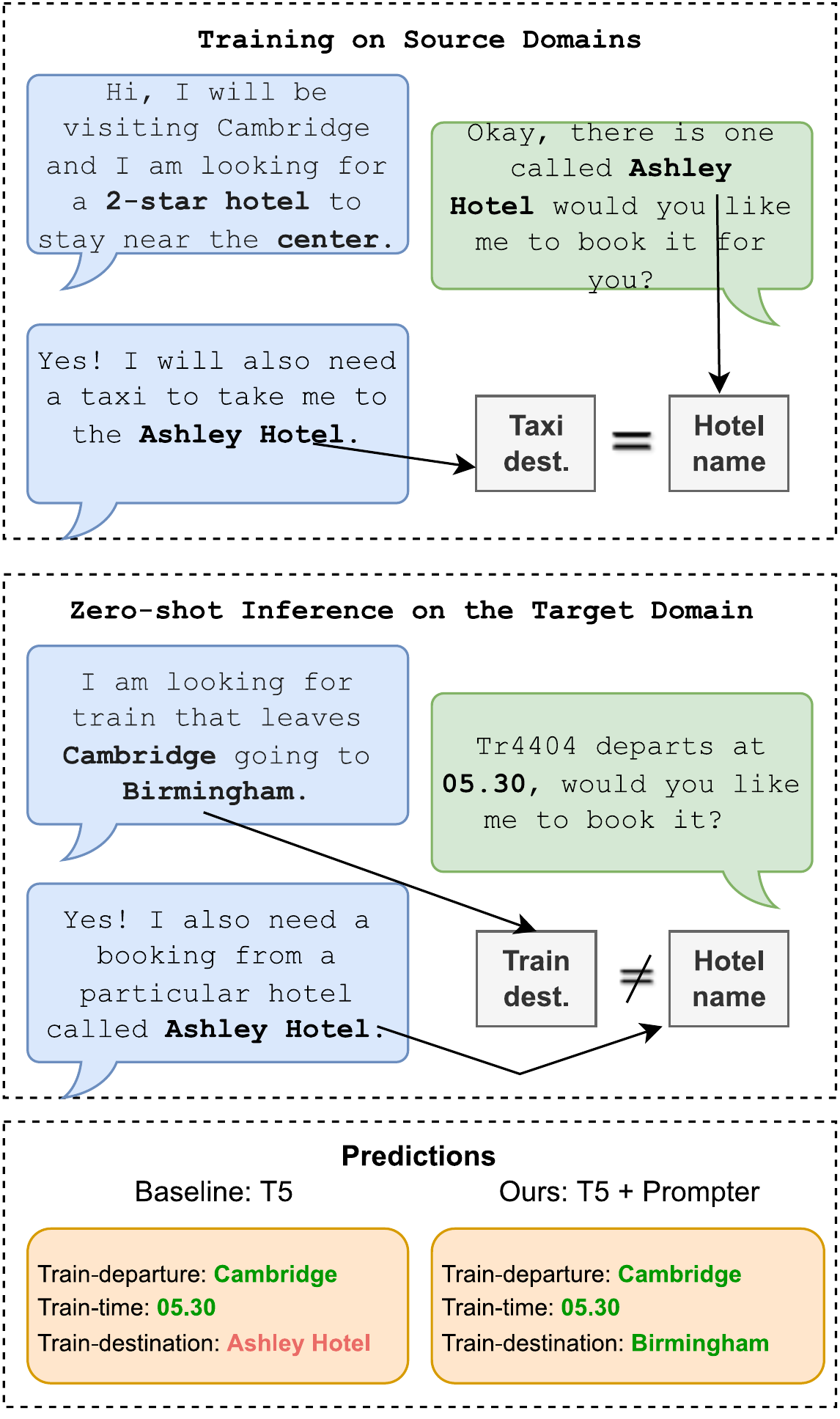}
\caption{Zero-shot domain adaptation. The model is trained on four source domains and tested on the train-booking domain without any supervised training. Bottom-left: T5 baseline predictions, Bottom-right: Prompter predictions. (Correct, incorrect) predictions are colored (green, red), respectively.
}
\label{Fig_intro}
\end{figure}

Task-oriented dialogue (TOD) systems serve users through several tasks, such as booking a table in a restaurant or suggesting tourist attractions. One crucial component of these systems, Dialogue State Tracking (DST),
is responsible for extracting users' preferences (\textit{i.e.} slot-values) over key attributes (\textit{i.e.} slot-labels) of their service~\cite{wu-etal-2019-transferable}.

DST has a significant role in TOD systems as it ensures that both the action taken in the back-end and the responses returned to the users are aligned with the preferences that the users indicate.

 A challenging task in this field is to adapt an existing DST model to a new domain it has not seen before without using any supervised data, \textit{i.e.} in the zero-shot scenario. This is important, as in many new scenarios, it is hard to collect data, let alone annotate it.  Yet it is still an essential need for a TOD system to appropriately answer such queries in new contexts. 
 The challenge arises from the differences in dialogue context, slot values, and slot labels among different domains.
 For example, a model could be trained on the `taxi-booking' domain and thus capable of extracting the destination for a taxi; but when deployed to the `train-booking' domain, the range of slot-values changes, resulting in a higher probability of a mistaken inference. We show an example (\Cref{Fig_intro}), where due to the superficial connections a baseline T5 model forms, it incorrectly predicts `Ashley Hotel' as the train destination (bottom left). 
 In many dialogue contexts, a large number of slots are unspecified. These are known as ``none''-valued slots. In cases where the model is adapting to a new domain without any prior training, it often incorrectly predicts none values. This makes it even more important to address the problem of domain shift.

 ~\citet{lin-etal-2021-leveraging} proposed to address this domain shift challenge via the language model's intrinsic ability to reason over prompts. Specifically, they concatenate the description of each slot as a hard prompt into the dialogue context and then generate the answers using the T5 model. While it does well for a naive baseline, it makes mistakes due to its superficial understanding of slot labels.

Meanwhile, another line of study has shown that Parameter-efficient Transfer Learning (PETL) methods are effective training methods to address domain shift. Due to the small number of parameters it introduces per task/instance, it overcomes  overfitting in  few-shot scenarios, outperforming earlier baselines.
There have been various attempts to use these methods for DST tasks within a few-shot, continual learning setting~\cite{zhu-etal-2022-continual, madotto-etal-2021-continual}. However, a significant barrier to adopting PETL is that such methods cannot be directly applied in  zero-shot, as they all require some form of supervised training.

In this study, we propose a new method to use prefix-tuning under a zero-shot scenario to benefit from the gains it brings for robustness,

even without supervised data. Rather than fine-tuning the prefixes during training, we add a new mechanism into the T5 architecture called Prompter\footnote{Implementation available at \url{ https://github.com/cuthalionn/Prompter}}. Prompter simply takes the description of the slot and then generates the prefixes on the fly. We then append these prefixes at each layer of the encoder to represent the dialogue from the perspective of the subject slot label.
This method makes minimal changes to LM parameters while generating unsupervised prefixes. This ensures both the preservation of general-purpose traits and extrapolation to new domains.

 We conduct experiments with the MultiWOZ 2.1 and SGD datasets.

 Prompter improves average JGA results across domains by 1.7 for MultiWOZ, and 9.1 points for the SGD dataset (considering 4 domains reported in prior studies) compared to the strongest baseline. This shows that PETL methods' robustness advantage is also favorable for unsupervised domain adaptation scenarios. To the best of our knowledge, these are the highest results achieved so far using a small language model.

Through further analysis, we have discovered that Prompter not only considers the semantic similarities of slot descriptions but also the frequencies in which slots co-appear in the dialogue context.
Furthermore, Prompter proves to be more effective in identifying slots that have no value within a conversation in comparison to previous methods.

\section{Related Work}
\paragraph{Dialogue State Tracking.}
DST has a long history of models working with a static, ontology-based problem definition (\textit{i.e.} slot-values are fixed)~\cite{balaraman-etal-2021-recent}. The static-ontology DST is a simplified classification problem where the model selects a value from each slot's value pool.~\cite{zhang-etal-2020-find,lee-etal-2019-sumbt,Rastogi2017ScalableMD, zhong-etal-2018-global}. Recently interest in {\it dynamic} ontologies have received attention, adding flexibility at inference time~\cite{wu-etal-2019-transferable,Rastogi2019TowardsSM, heck-etal-2020-trippy}.
\paragraph{Low-resource Domain Adaptation.}
 Dynamic ontology introduces slot-value level flexibility, but its ability to work with new slot-labels is limited. Domain adaptation of DST systems aims to make the model adaptable to new domains/slot-labels. Few studies have attempted to utilize language models' intrinsic reasoning abilities by mapping DST as a question--answering task

 ~\cite{lin-etal-2020-mintl,Zhou2019MultidomainDS}. \citet{shin-etal-2022-dialogue}, on the other hand, map DST to a dialogue summarization task, and \citet{Xie2022UnifiedSKGUA} map it to a structured-knowledge grounding task. Many use data augmentation to address the lack of supervision in the target domain~\cite{Qiu2022StructureEI,mi-etal-2021-self,gritta-etal-2021-conversation,aksu-etal-2022-n,DBLP:journals/corr/abs-2010-12850}. Finally, remaining studies focus on improving the model's architecture and training strategies for robustness toward domain changes.~\cite{feng-etal-2022-dynamic,Balaraman2020DomainAwareDS, Madotto2020LanguageMA,huang-etal-2020-meta,coope-etal-2020-span,wu-etal-2019-transferable,lei-etal-2018-sequicity,lin-etal-2021-leveraging,Yang2022PromptLF}. ~\citet{wang-etal-2022-slot} have a similar goal as our own, but they use a different method. They create cross-slot dependency by combining multiple slot prompts to create a final prompt, which encourages the model to apply what it has learned in one slot to other slots.

\paragraph{PETL for DST Domain Adaptation.}
Parameter Efficient Transfer Learning (PETL) is a recently trending set of methods that aims to adapt models more efficiently by significantly reducing the number of parameters that need to be fine-tuned.~\cite{pfeiffer-etal-2020-adapterhub,lester-etal-2021-power,liu-etal-2022-p,li-liang-2021-prefix,Houlsby2019ParameterEfficientTL}. 
Many studies have found that PETL is advantageous for low-resource domain adaptation settings due to its efficient parameter training scheme. This scheme minimizes changes in LM parameters and thus believed to prevent over-fitting~\cite{li-liang-2021-prefix,liu-etal-2022-p}. However,~\citet{He2022HyperPromptPT} argues that tuning the entire language model does not negatively impact its robustness advantage. Researchers in the DST field have also utilized PETL methods for their robust capabilities. In their work,~\citet{zhu-etal-2022-continual} employed soft prompts and fine-tuned them for each domain in a continual learning setting, utilizing validation sets from target domains to decide which previous prompts to use for initialization.~\citet{madotto-etal-2021-continual} also tackled the problem of continual learning, using unique adapters for each domain and relying on a classifier to select which adapter to use during inference. Both studies only explored the use of PETL methods for DST with few-shot availability. In contrast, this study aims to investigate a well-known PETL method, prefix-tuning~\cite{li-liang-2021-prefix}, for zero-shot domain adaptation of DST models.
 
\section{Background}

 \begin{figure*}[htb!]
\centering
\includegraphics[width=\textwidth]{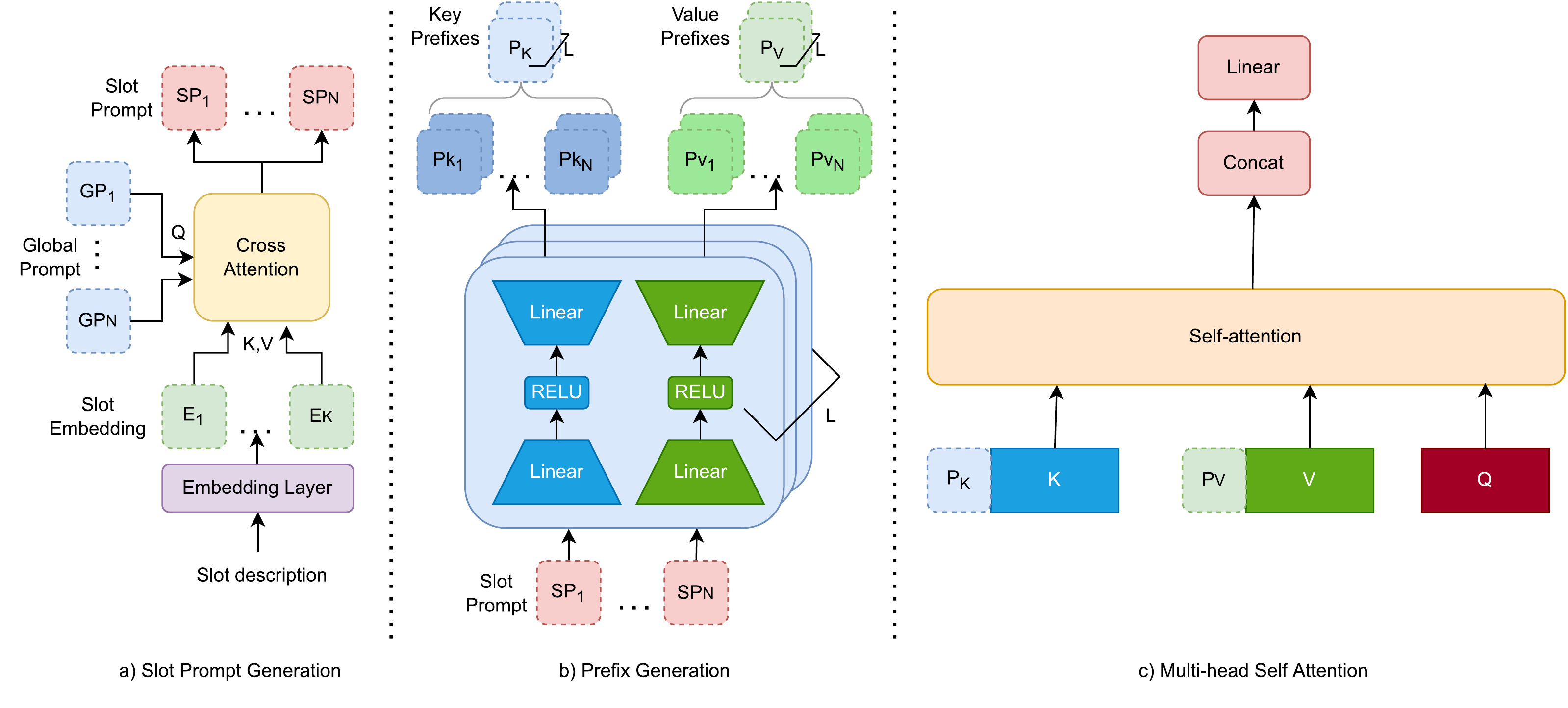}
\caption{The architecture of our proposed method, Prompter. Prompter leverages the prefix-tuning method to enable zero-shot learning without the need for supervised data and it is composed of three parts: (a) Slot Prompt Generation where the information from the description is fused with some global prompt to generate slot-specific prompts, (b) Prefix Generation which feeds slot prompts across two linear layers and an activation function to generate per-layer key and value prefixes, (c) Finally these prefixes are concatenated to keys and values at every layer of the T5 encoder.}
\label{Fig:Method}
\end{figure*}
\subsection{Dialogue State Tracking Task}
 A task-oriented dialogue consists of a number of consecutive system and user utterances, together referred to as a turn, $t_i = (s_i,u_i)$. Each turn is annotated with a belief state that shows the user's preferences over a number of attributes from various domains up to and including that turn, $B_i=(D_0, D_1, ..., D_K)$ where $D_j$ is the belief state for domain $j$, and $K$ is the total number of domains. The belief state for each domain is made up of a list of slot-label (\textit{e.g.} `restaurant-area') and slot-value pairs (\textit{e.g.} `center'), $D_j=\{s_0:v_0,s_1:v_1, ..., s_N:v_N\}$, where $N$ is the number of slots within domain $j$. Each $s_i$ is further annotated with a description that explains the attribute in the context of the domain (\textit{e.g.} `restaurant-area':`The area of the city where the restaurant is located.'). For each $v_i$, if $s_i$ is not discussed in the dialogue context, $v_i$ is set to `none'. Otherwise, $v_i$ is a sequence of tokens. The task of DST is to predict the belief state $B_i$ for a given dialogue context $DC$, \textit{i.e.} dialogue turn history up to and including turn $i$, $DC = (t_0,t_1, ..., t_i)$. 
\subsection{Prefix-Tuning}
Prefix-tuning is a parameter-efficient alternative to fine-tuning which optimizes a small continuous task-specific vector called the prefix for each new task. These tunable prefix vectors are prepended to the keys and values of the multi-head attention at every layer of the transformer~\cite{li-liang-2021-prefix,He2021TowardsAU}. \citet{li-liang-2021-prefix} also report that 
prefix-tuning also improves extrapolation to unseen tasks in few-shot settings. However, there is no straightforward way to use this method for the zero-shot setting, as it requires supervision to fine-tune the prefixes.
\section{Method}
 
We propose to add a new mechanism into the T5 architecture~\cite{Raffel2019ExploringTL}, called Prompter, to take advantage of prefix-tuning's extrapolation capabilities without requiring supervision. Instead of fine-tuning the prefixes with source domain data, we generate them on the fly for each slot. However, we need a way to condition Prompter for a new domain without any supervised data. Task-oriented dialogue schemas provide a solution by annotating the slot descriptions for each slot-label. Using these slot descriptions Prompter can generate domain-specific prefixes which allow it to adapt to any domain without the need for supervised data. We can summarize the Prompter pipeline in three key parts: (1) Slot Prompt Generation, (2) Prefix Generation, and (3) Multi-head Self Attention.

\paragraph{Slot Prompt Generation.}
is responsible for generating a prompt that is specific to each slot, using its unique description. Previous approaches to this problem, such as simply concatenating the description to the input, result in only a superficial understanding of the slots in zero-shot settings~\cite{lin-etal-2021-leveraging}. Additionally, using slot embeddings as soft prompts can cause unstable training and hinder zero-shot adaptation due to changes in the descriptions.  Instead, we propose using a global prompt that is modified according to each slot's description. This modification is applied through a cross-attention mechanism that attends the global prompt to the slot description's embedding, \textit{c.f.} Figure~\ref{Fig:Method}a. This approach ensures that each slot prompt shares the same initialization addressing unstable training, and the modifications reflect changes in the slot-label addressing domain shift. It also has the advantage of making the final prompt's length fixed, regardless of the length of the description. The slot prompt is calculated as follows:
\begin{equation}
   S =  ((GW_q) (EW_k)^\top ) (EW_v)  
\end{equation}
where $W_q$,$W_k$, and $W_v \in \mathbb{R}^{d \times d} $  are query, key, and value weights for the cross attention mechanism and d is the model dimension, $G \in \mathbb{R}^{N\times d}$ is the global prompt\footnote{For N we try different values from [1,100] range and empirically found 10 to work best. Thus we set N=10 throughout conducted experiments.}, $E \in \mathbb{R}^{K \times d}$ is the slot embedding, $K$ is the length of slot description, and $S \in \mathbb{R}^{N\times d}$ is the slot prompt.

\paragraph{Prefix generation.}
For the DST task, the dialogue context can make up the majority of the language model input (\textit{i.e.} 100--400 tokens long dialogue context compared to 10--15 tokens long slot description), this results in challenges with the prompt-tuning method because the prompt's impact can vanish easily before the decoding starts. This is why we opt for prefix-tuning because it ingests prompts at each layer and thus the generated value will have higher exposure to the prompt.

So following the generation of slot prompts the next step is to generate key and value prefixes for each layer. For this step, we have tried several different architectural designs such as a simple MLP or a whole transformer block. We empirically observed that while the former lags behind due to the small number of parameters the latter results in overfitting. Thus, inspired by~\citet{He2022HyperPromptPT} we use a sequence of down and up projections separated by an activation function as prefix generators, \textit{c.f.}~Figure\ref{Fig:Method}b. Note that each transformer layer has a pair of dedicated prefix generators to generate key and value prefixes:
\begin{equation}
    K_i = RELU(SWk_{down_i}){Wk}_{up_i}
\end{equation}
\begin{equation}
    V_i = RELU(SWv_{down_i}){Wv}_{up_i}
\end{equation}
where $K_i$, and $V_i \in \mathbb{R}^{N\times d}$ are key and value prefixes for the $i^{th}$ layer; ${{Wk}_{down_i}}$, ${Wv}_{down_i} \in \mathbb{R}^{d\times r} $, ${Wk}_{up_i}$ and ${Wv}_{up_i} \in \mathbb{R}^{r\times d}$ are the respective down and up projectors for the $i^{th}$ layer; $r$ is the bottleneck dimension.  $r$ is set to $d/4$ throughout our experiments.

\paragraph{Multi-head Self Attention.}
After we get $K_i$ and $V_i$ for each layer $i$ we split them to $N_{h}$ head vectors $K_i^{j}$ and $V_i^{j} \in \mathbb{R}^{N\times d_h}$ for each head $j$, where $d_h = d / N_h$ is the dimension per head. Finally, we concatenate these key and value prefixes into the self-attention mechanism at each layer of the transformer encoder completing our modifications to the original T5 architecture, \textit{c.f.}~Figure~\ref{Fig:Method}c.

\begin{equation}
     head_i^{j} = ( h_iW_{q_i}^{j}
     [K_i^{j},h_iW_{k_i}^{j}]^\top)[V_i^{j},h_iW_{v_i}^{j}]
\end{equation}
where $head_i^{j}$ is the output from the $j^{th}$ head of self-attention mechanism at layer $i$; $W_{q_i}^{j}$, $W_{k_i}^{j}$, and $W_{v_i}^{j} \in \mathbb{R}^{d\times d_h}$ are query, key and value weight matrices of the $j^{th}$ head in the $i$th layer; and $h_i$ is the input to the $i^{th}$ layer.

The final output of the multi-head self-attention at layer $i$ is calculated as:
\begin{equation}
    MSA(h,i) = [head_i^{0},head_i^{1}, ..., head_i^{N_h}]W_{o_i} 
\end{equation}
where $W_{o_i} \in \mathbb{R}^{d\times d}$.


\begin{table*}[!ht]
\centering
\begin{tabular}{|c|c|c|c|c|c|c|c|}
\hline
\begin{tabular}[c]{@{}l@{}}Model\end{tabular} & \begin{tabular}[c]{@{}l@{}}Lang.\\ Model\end{tabular} & Attraction    & Hotel          & Restaurant     & Taxi          & Train       & Avg           \\ \hline
TRADE                                                      & -                                                        & 20.06         & 14.20          & 12.59          & 59.21         & 22.39       & 25.69         \\ \hline
MA-DST                                                     & -                                                        & 22.46         & 16.28          & 13.56          & 59.27         & 22.76       & 26.87         \\ \hline
SUMBT                                                      & BERT-b                                                & 22.60         & 19.08          & 16.50          & 59.50         & 22.50       & 28.18         \\ \hline
~\citeauthor{li-etal-2021-zero}                                           & GPT2                                                  & 23.67       & 18.54 & 21.05 & 59.1         & 24.34       & 29.34        \\ \hline
T5DST                                                      & T5-s                                                 & 31.92         & \textbf{20.72 }         & 20.09          & 64.12         & 28.83       & 33.56         \\ \hline
~\citeauthor{wang-etal-2022-slot}                                         & T5-s                                                 & 33.92         & 18.85          & 20.75          & 66.25         & 36.96       & 35.55         \\ \hline
T5DST$^{*}$                                                      & PPTOD-s            & $35.5_{\pm1.7}$         & $20_{\pm0.9}$             & $25.3_{\pm0.8}$           & $65.6_{\pm 0.6}$          & $35.3_{\pm1.0}$        & $36.4_{\pm6.9}$          \\ \hline
Prompter$^{*}$                                          & PPTOD-s             & $\textbf{35.8}_{\pm0.7}$ & $19.2_{\pm0.8}$           & $\textbf{26}_{\pm0.7}$             & $\textbf{66.3}_{\pm0.2}$ & $\textbf{39}_{\pm0.5}$ & $\textbf{37.2}_{\pm7}$ \\ \hline
\end{tabular}
\caption{ Zero-shot join-goal accuracy(\%) results on MultiWOZ 2.1 dataset. Results for all baselines are reported from original papers. Models with * trained using the semi-frozen training scheme. For our trained models the results are averaged over three runs. The best results on each column are \textbf{bold}.}
\label{Tab:MWOZ}
\end{table*}

\begin{table*}[!ht]
\centering
\begin{tabular}{|c|c|c|c|c|c|c|c|}
\hline
JGA                                                       & Buses           & Messaging          & Trains & Payment          & Media      & Events           & Unseen        \\ \hline
SGD-baseline   & 9.7           & 10.2   &   13.6    & 11.5          & 18.0        & 23.5        & -           \\ \hline
Seq2seq-DU     & 16.8          & 4.9    &   16.8    & 7.2           &     -    & -       & -           \\ \hline
Transfer-QA           & 15.9          & 13.3  &   17.4         &  \textbf{24.7}        &   -   & -         & -           \\ \hline
~\citeauthor{wang-etal-2022-slot}                                              & 43.9          & 36.6      & 46.7    & 16.5         & -     & -         & -           \\ \hline
T5DST$^{*}$                                                     & $46.8_{\pm2.2}$          & $54_{\pm2.8}$  &   $\textbf{53}_{\pm0.4}$     & $23.3_{\pm3.8}$ &  $55.5_{\pm3.3}$   &      $48.8_{\pm2.5}$    & $48.0_{\pm0.8}$          \\ \hline
Prompter$^{*}$ & $\textbf{48.4}_{\pm2.1}$ & $\textbf{59.2}_{\pm1.3}$ & $50.8_{\pm0.9}$   & $21.9_{\pm4.6}$          &   $\textbf{65.3}_{\pm3.8}$    &  $\textbf{51.5}_{\pm0.4}$ & $\textbf{49.4}_{\pm0.4}$ \\ \hline
\end{tabular}
\caption{Zero-shot joint-goal accuracy (\%) results on SGD dataset. Results for all baselines are reported from original papers. Models with * trained using the semi-frozen training scheme. For our trained models the results are averaged over three runs. The final column shows the average JGA on all unseen slots. The best results on each column are \textbf{bold}.}
\label{Tab:SGD}
\end{table*}
\section{Experimental Setup}
\subsection{Datasets}
 We conduct experiments with two well-known DST benchmarks: MultiWOZ and SGD~\cite{budzianowski-etal-2018-multiwoz,Rastogi2019TowardsSM}. MultiWOZ is a task-oriented dialogue dataset collected in a wizard of oz setting using human speakers. It has 10k dialogues that span over 7 domains. It provides turn-level annotations and descriptions of each slot label. In line with previous studies, we limited our experiments to only 5 domains because the police and hospital domains do not have a sufficient number of examples in the test set. We use MultiWOZ version 2.1 which addresses the noisy state annotations within the original dataset~\cite{eric-etal-2020-multiwoz}. Similar to MultiWOZ, the SGD dataset also has turn-level annotations and descriptions, \textit{i.e.} schema, for each domain and slot. It has over 20k annotated conversations between a human and a virtual assistant. These span over 20 domains. Besides, the SGD dataset has unseen domains in the test set specifically formed to evaluate zero-shot performance.
\subsection{Baseline Models}
We compare our method with a range of DST models from the past as well as the recent state of the art. The only models we utilize that do not depend on a language model are \textbf{TRADE}~\cite{wu-etal-2019-transferable} and \textbf{MA-DST}~\cite{Kumar2020MADSTMB}. The former introduces the copy mechanism to ease predicting slots not seen during training, whereas the latter adds cross-attention to model relationships between the context and slots at different semantic levels and self-attention to resolve cross-domain coreferences to a base RNN layer. \textbf{SUMBT} by ~\citet{lee-etal-2019-sumbt} is built with BERT and again uses an attention mechanism to learn relations between domains and slots. \textbf{SGD-baseline}~\cite{Rastogi2019TowardsSM} feeds slots, domains, and value embeddings into a BERT encoder to create schema embedding and uses it to predict dialog state in the target domain under zero-shot. \textbf{Seq2seq-DU}~\cite{feng-etal-2021-sequence} formalizes DST as a sequence-to-sequence task where the dialog history is transformed directly into semantic frames.~\citet{li-etal-2021-zero} on the other hand use GPT-2 and define DST as a generative question-answering approach.  \textbf{TransferQA} builds on a similar motivation but combines both extractive and multi-choice QA enabling tracking categorical and non-categorical slots simultaneously~\cite{lin-etal-2021-zero}. \textbf{T5DST}~\cite{lin-etal-2021-leveraging} and \textbf{~\citet{wang-etal-2022-slot}} both use the T5 architecture. The former concatenates slot descriptions with dialogue context and generates slot values in an auto-regressive manner. Whereas the latter proposes a unique design that models cross-slot dependency by composing multiple slots as the final prompt so that the model is forced to learn the relations among each slot.

\subsection{Training Details}
\label{Sec:Training}
For all experiments, we used a Tesla-V100 GPU. We use the small-sized PPTOD~\cite{su-etal-2022-multi} built on the T5 architecture for the T5DST baseline and our own Prompter. We empirically found PPTOD to be more suitable for prompt-tuning tasks most probably due to the nature of its pretraining tasks. We set the batch size to 8 with gradient accumulation every 8 steps. We use AdamW optimizer~\cite{Loshchilov2017FixingWD} for training and set the initial learning rate to $1e-4$.
\paragraph{Semi-frozen Training Scheme}Contrary to what is typically recommended for limited data scenarios by traditional PETL techniques, we discovered that freezing LM parameters does not improve performance in the zero-shot scenario. This is in line with what \citet{He2022HyperPromptPT} suggests. However, we also find that tuning all parameters is imperfect.
In search for a better strategy we experiment with different combinations of frozen layers and compare the results for zero-shot train domain performance. We found that the best strategy is a semi-frozen (S.F.) training scheme, where all LM parameters are trained for 1k steps and then all layers of the T5 model are frozen except the first and last layers of the encoder and decoder (\textit{c.f.} \Cref{sec:semi_frozen_search} for more details). Thus for the experiments conducted in this section, we employ this strategy to train the models.
\subsection{Evaluation}
We evaluate the performance of all models using Joint Goal Accuracy (JGA) following prior studies. For MultiWOZ, a zero-shot setting is used where training occurs on four domains and the remaining domain is used for testing. For SGD, results are reported on domains that are not included in both the training and validation sets, as they have already been included in the PPTOD pretraining. We modified the official SGD evaluation script to reflect this change. Therefore, in our evaluation settings, unseen domains refer only to domains in the test data, contrary to the original definition by~\citet{Rastogi2019TowardsSM} which considers domains only showing up in the validation data unseen as well.

\section{Results and Analysis}
\label{Sec:res}

In MultiWOZ (Table~\ref{Tab:MWOZ}), our addition of Prompter shows improvements in all domains except Hotel, boosting the average JGA by 1.7 points, compared to the state-of-the-art model by~\citet{wang-etal-2022-slot}. We believe the lack of improvements in the hotel domain for Prompter is due to it having many unique slots (\textit{i.e.} `hotel-internet', `hotel-parking', `hotel-type', \textit{etc.}). This makes it harder to take advantage of earlier domains as they lack similar slots. This is also in line with the results from~\citet{wang-etal-2022-slot}, as their cross-slot dependency design also lags behind for hotel domain results.

We also present the results on the SGD dataset in Table~\ref{Tab:SGD}, where Prompter shows improvements on average. We share results over 6 representative domains along with results for official unseen domain performance.

Once more, Prompter demonstrates superior performance on average in unfamiliar domains. Compared to the results reported in the original paper by \citet{wang-etal-2022-slot} for four domains (Columns 1 through 4 of Table \Cref{Tab:SGD}), Prompter shows an average improvement of $9.1$ in JGA. The Alarm domain is excluded from the comparison as PPTOD has been pretrained on it.

\subsection{Ablation Study}
\label{Sec:ablation}

We further conducted ablation to analyze the contribution of Prompter's components (Table~\ref{Tab:ablation}). 

Adding the S.F. training scheme (second row) to the T5DST baseline introduces performance increase across all domains. This demonstrates that this training scheme plays a significant role in the robustness of the model. If we switch the pre-trained model from T5 to PPTOD (third row), we see another round of improvement but it is inconsistent across domains. 

Finally, it is evident from the final row that adding the Prompter increases the results by another margin, clearly showing its contribution.
\begin{table}
\small
\begin{tabular}{|cll|c|c|c|c|c|}
\hline
\multicolumn{3}{|c|}{Model}        & Train       & Rest        & Hotel         & Taxi          & Attr          \\ \hline
\multicolumn{3}{|c|}{T5DST}     & 28.83       & 20.09       & 20.72         & 64.12         & 31.92         \\ \hline
\multicolumn{3}{|c|}{+ S.F.}    & 29.3        & 24.4        & \textbf{22.3} & 65.6          & 34.76         \\ \hline
\multicolumn{3}{|c|}{+ PPTOD} & 35.3        & 25.3        & 20            & 65.6          & 35.5          \\ \hline
\multicolumn{3}{|c|}{+ Prompter} & \textbf{39} & \textbf{26} & 19.2          & \textbf{66.3} & \textbf{35.8} \\ \hline
\end{tabular}
\caption{Ablation results on the test set of MultiWOZ 2.1. We cumulatively add semi-frozen (S.F.) training, PPTOD, and Prompter  to the T5DST baseline and report results. The best results along each column are \textbf{bold}.}
\label{Tab:ablation}
\end{table}

\begin{table*}
\resizebox{\textwidth}{!}{%
\begin{tabular}{|l|lll|lll|lll|lll|lll|}
\hline
\multirow{2}{*}{}                                         & \multicolumn{3}{l|}{Attraction}                                                                                                 & \multicolumn{3}{l|}{Hotel}                                                                                                      & \multicolumn{3}{l|}{Restaurant}                                                                                                 & \multicolumn{3}{l|}{Taxi}                                                                                                      & \multicolumn{3}{l|}{Train}                                                                                                    \\ \cline{2-16} 
                                                          & \multicolumn{1}{l|}{MP↓}            & \multicolumn{1}{l|}{OP↓}            & \begin{tabular}[c]{@{}l@{}}None\\ Acc↑\end{tabular} & \multicolumn{1}{l|}{MP↓}            & \multicolumn{1}{l|}{OP↓}            & \begin{tabular}[c]{@{}l@{}}None\\ Acc↑\end{tabular} & \multicolumn{1}{l|}{MP↓}            & \multicolumn{1}{l|}{OP↓}            & \begin{tabular}[c]{@{}l@{}}None\\ Acc↑\end{tabular} & \multicolumn{1}{l|}{MP↓}            & \multicolumn{1}{l|}{OP↓}           & \begin{tabular}[c]{@{}l@{}}None\\ Acc↑\end{tabular} & \multicolumn{1}{l|}{MP↓}           & \multicolumn{1}{l|}{OP↓}           & \begin{tabular}[c]{@{}l@{}}None\\ Acc↑\end{tabular} \\ \hline
T5DST                                                     & \multicolumn{1}{l|}{76.58}          & \multicolumn{1}{l|}{13.78}          & 65.96                                               & \multicolumn{1}{l|}{\textbf{59.51}} & \multicolumn{1}{l|}{24.03}          & \textbf{76.30}                                      & \multicolumn{1}{l|}{35.98}          & \multicolumn{1}{l|}{14.25}          & 79.15                                               & \multicolumn{1}{l|}{\textbf{53.57}} & \multicolumn{1}{l|}{10.96}         & 83.94                                               & \multicolumn{1}{l|}{\textbf{3.93}} & \multicolumn{1}{l|}{12.81}         & 90.91                                               \\ \hline
\begin{tabular}[c]{@{}l@{}}T5DST+\\ Prompter\end{tabular} & \multicolumn{1}{l|}{\textbf{69.80}} & \multicolumn{1}{l|}{\textbf{11.16}} & \textbf{68.81}                                      & \multicolumn{1}{l|}{65.77}          & \multicolumn{1}{l|}{\textbf{21.66}} & 75.34                                               & \multicolumn{1}{l|}{\textbf{29.14}} & \multicolumn{1}{l|}{\textbf{13.40}} & \textbf{82.07}                                      & \multicolumn{1}{l|}{54.40}          & \multicolumn{1}{l|}{\textbf{8.92}} & \textbf{84.05}                                      & \multicolumn{1}{l|}{7.66}          & \multicolumn{1}{l|}{\textbf{9.08}} & \textbf{91.50}                                      \\ \hline
\end{tabular}%
}
\caption{Fine Grained Analysis over MultiWOZ 2.1 dataset. MP and OP stand for miss-prediction and over-prediction respectively. Down arrow (↓) means lower the better, whereas up arrow (↑) means higher the better. The best results among each column are \textbf{bold}.}
\label{Tab:finegrained}
\end{table*}
\subsection{Fine Grained Analysis}
\label{Sec:fineg}

\paragraph{How does Prompter improve results?}We define two new metrics to better understand  Prompter's improvements: {\it Miss-prediction} (MP), where the  model fails to correctly identify a gold slot-label, mistakenly labeling it as `none' instead; and {\it Over-prediction} (OP), where the model incorrectly predicts a `none' valued slot-label as something else. We then combine these metrics in {\it None Accuracy}, a metric that measures the accuracy of the model's predictions regarding the ``activeness'' of a slot-label. In other words, it measures how often the model correctly predicts whether a slot-label has the value `none' or not. The results over all 5 domains can be found in Table~\ref{Tab:finegrained}. It is evident that Prompter's improvement comes from the None accuracy measure as its results are in line with the change in JGA (\textit{i.e.} improvements across all domains except the Hotel domain). Moreover, we find that this is mostly due to the reduction of over-prediction mistakes --- Prompter decreases this class of error in every domain. 

\begin{figure}[!ht]
     \centering
     \hfill
          \begin{subfigure}[b]{0.48\textwidth}
         \centering
         \includegraphics[width=\textwidth]{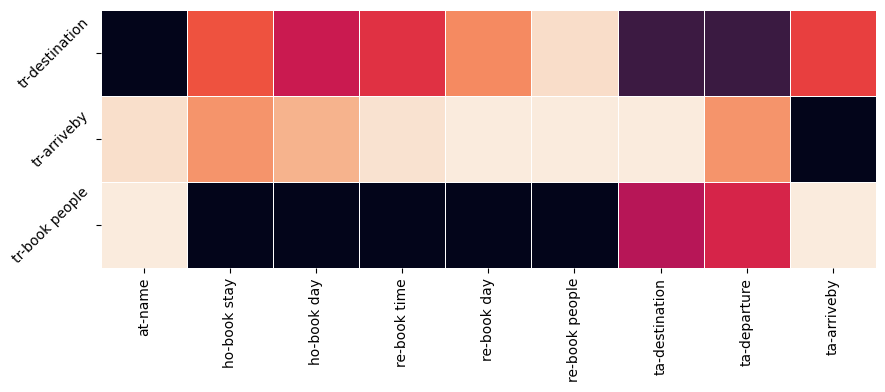}
         \caption{Source domain slots in close proximity to `Train-destination', `Train-arriveby', and `Train-bookpeople' slots, according to generated prefix similarities.}
         \label{subfig:train_heat}
     \end{subfigure}
     \hfill
             \begin{subfigure}[b]{0.48\textwidth}
         \centering
         \includegraphics[width=\textwidth]{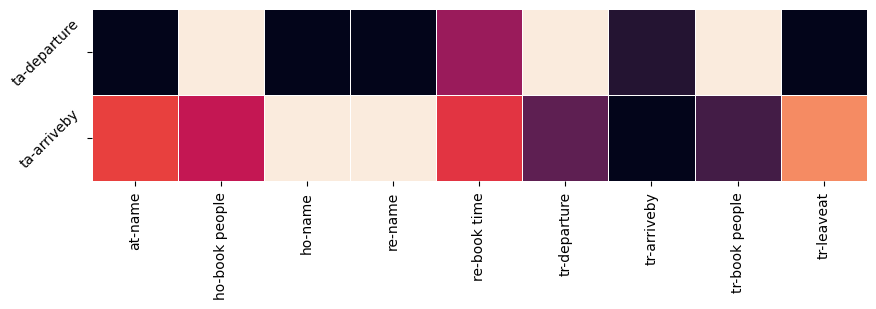}
         \caption{Source domain slots close to `Taxi-departure' and `Taxi-arriveby' slots, according to generated prefix similarities.}
         \label{subfig:taxi_heat}
     \end{subfigure}
             \caption{Heatmaps depicting the similarity of selected source and target domain slots. The generated prefixes are aggregated and compared with cosine similarity, where darker colors indicate higher similarity.}
        \label{fig:heatmap}
\end{figure}
\begin{table*}[!ht]
\resizebox{\textwidth}{!}{%
\begin{tabular}{|c|ll|}
\hline
Dial                                     & \multicolumn{2}{l|}{Conversation Details}                                                                                                                                                                             \\ \hline
\multirow{10}{*}{1}                      & \multicolumn{1}{l|}{U0}                                                        & I am looking for a train that leaves on Wednesday, going to Bishops Stortford.                                                       \\
                                         & \multicolumn{1}{l|}{S1}                                                        & tr4404 departs \textbf{Cambridge} at 05:29 and arrives at 06:07 in Bishops Stortford. Would you like me to book it?                          \\
                                         & \multicolumn{1}{l|}{U1}                                                        & Thats pretty early. Would there be a later train that arrives by 15:00?                                                             \\
                                         & \multicolumn{1}{l|}{S2}                                                        & tr3844 departs Cambridge Wednesday at 13:29 and arrives in Bishops Stortford by 14:07. Would you like to reserve a seat?             \\
                                         & \multicolumn{1}{l|}{U2}                                                        & No. How long will the train take?                                                                                                    \\
                                         & \multicolumn{1}{l|}{S3}                                                        & That train ride will take approximately 38 minutes.                                                                                  \\
                                         & \multicolumn{1}{l|}{U3}                                                        & Thanks. I also need a particular hotel. Its name is \textbf{Ashley hotel}.                                                                    \\ \cline{2-3} 
                                         & \multicolumn{1}{l|}{GT}                                                        & \{train-destination: Bishops Stortford, train-day: Wednesday, train-arriveby: 15:00, train-departure: \textbf{Cambridge}\}    \\ \cline{2-3} 
                                         & \multicolumn{1}{l|}{T5DST}                                                     & \{train-destination: Bishops Stortford, train-day: Wednesday, train-departure: \textbf{Ashley hotel}, train-arriveby: 15:00\} \\ \cline{2-3} 
                                         & \multicolumn{1}{l|}{\begin{tabular}[c]{@{}l@{}}T5DST+\\ Prompter\end{tabular}} & \{train-destination: Bishops Stortford, train-day: Wednesday, train-arriveby: 15:00, train-departure: \textbf{Cambridge}\}    \\ \hline

\multicolumn{1}{|l|}{\multirow{7}{*}{2}} & \multicolumn{1}{l|}{U0}                                                        & I am coming to Cambridge and would like to see some \textbf{architecture}. Do you have any located in the \textbf{centre}?                             \\ \cline{2-2}
\multicolumn{1}{|l|}{}                   & \multicolumn{1}{l|}{S1}                                                        & Yes, there are 5 places located in the centre. I recommend the All Saint Church on Jesus Lane.                                       \\ \cline{2-2}
\multicolumn{1}{|l|}{}                   & \multicolumn{1}{l|}{U1}                                                        & Thanks! What is the entrance fee?                                                                                                    \\ \cline{2-2}
\multicolumn{1}{|l|}{}                   & \multicolumn{1}{l|}{S2}                                                        & ...                                                                                                                                  \\ \cline{2-3} 
\multicolumn{1}{|l|}{}                   & \multicolumn{1}{l|}{GT}                                                        & \{\}                                                                                                                                 \\ \cline{2-3} 
\multicolumn{1}{|l|}{}                   & \multicolumn{1}{l|}{T5DST}                                                     & \{hotel-type: \textbf{architecture}, hotel-area: \textbf{centre}\}                                                                             \\ \cline{2-3} 
\multicolumn{1}{|l|}{}                   & \multicolumn{1}{l|}{\begin{tabular}[c]{@{}l@{}}T5DST+\\ Prompter\end{tabular}} & \{\}                                                                                                                                 \\ \hline

\multicolumn{1}{|l|}{
\multirow{8}{*}{3}} & \multicolumn{1}{l|}{U0}                                                        & Hello, I am looking for places to go in the centre?                             \\ \cline{2-2}
\multicolumn{1}{|l|}{}                   & \multicolumn{1}{l|}{S1}                                                        & There are many attractions in the centre like museums, architecture, boating, and concert halls. What are you interested in?                                       \\ \cline{2-2}
\multicolumn{1}{|l|}{}                   & \multicolumn{1}{l|}{U1}                                                        & How about a boating attraction?                                                                                                    \\ \cline{2-2}
\multicolumn{1}{|l|}{}                   & \multicolumn{1}{l|}{S2}                                                        & There are 2 in the centre of town. Scudamores punting co., and the cambridge punter. Would either of those interest you?                                                                                                                                  \\ \cline{2-2}
\multicolumn{1}{|l|}{}                   & \multicolumn{1}{l|}{U2}                                                        & Could you give me the address for the Cambridge punter, please? I also need a place to stay, preferably somewhere \textbf{cheap}.                                                                                                                                  \\ \cline{2-3} 
\multicolumn{1}{|l|}{}                   & \multicolumn{1}{l|}{GT}                                                        & \{hotel-pricerange: \textbf{cheap}\}                                                                                                                                 \\ \cline{2-3} 
\multicolumn{1}{|l|}{}                   & \multicolumn{1}{l|}{T5DST}                                                     & \{hotel-pricerange: cheap\}                                                                             \\ \cline{2-3} 
\multicolumn{1}{|l|}{}                   & \multicolumn{1}{l|}{\begin{tabular}[c]{@{}l@{}}T5DST+\\ Prompter\end{tabular}} & \{hotel-pricerange: \textbf{cheap}, hotel-type: \textbf{cheap}, hotel-internet: \textbf{cheap}\}                                                                                                                                 \\ \hline
\end{tabular}%
}
\caption{Three example dialogues from the MultiWOZ 2.1 test set. Each dialogue consists of user and system turns, ground truth dialogue state (GT).  We show a pair of predictions by the T5DST baseline, and our Prompter.}
\label{Tab:case}
\end{table*}

\paragraph{How does Prompter connect slots?}To better understand the benefits of using Prompter, we look at how it connects target domain slots with source domain slots. This is done by aggregating the key prefixes across each layer and attention head for every slot and then comparing them to the source domain slot prefixes from the training set using cosine similarity. 

Figure~\ref{fig:heatmap} highlights important similarities among some of the taxi and train domain slots (\textit{c.f.} Appendix~\ref{App:prefix} for a comprehensive version that includes all domains and slots).
Figure~\ref{subfig:train_heat} shows that `train-destination' has a high similarity with `taxi-departure' and `destination', as well as the `attraction-name' slots. The first two connections are expected, but the latter is also relevant because the `attraction-name' often appears as the `taxi-destination' in training.
This indicates that the model finds that the `destination' slots can often contain named entities (such as locations) within the dialogue. For `train-arriveby', the most similar slot is also the semantically closest: `taxi-arriveby'. Finally, for the `train-bookpeople' slot, the most similar slots are those related to booking from the hotel and restaurant domains, which makes sense as these often co-occur in the training data. 

Figure~\ref{subfig:taxi_heat} shows the results of adapting in the taxi domain. The similarity between the `taxi-arriveby' slot and its train domain counterpart, `train-arriveby', is high as expected. Moreover, for the `taxi-departure' slot, the generated prefixes are most similar to slots for attraction, restaurant, and hotel names. This is likely because the `train-departure' slot also has named entities as values. 

The findings show that Prompter not only utilizes slots with similar descriptions to create prefixes, but also accounts for other slots that co-occur in the same conversation with a similar source slot. This is important as slots may have different descriptions but exhibit significant semantic overlap (e.g., `taxi-departure' and `hotel-name' having location named entities as values).

\subsection{Case study}
We use three dialogues from the MultiWOZ test set to demonstrate some of the phenomena observed in previous analysis studies (Table~\ref{Tab:case}).
The first example shows how the T5DST baseline is susceptible to overgeneralization from training data. When the T5DST model encounters a hotel name during zero-shot inference on the train domain, it mistakenly assumes that the hotel is the departure for the train because it has been trained to associate location names with taxi departure/destination. Prompter avoids this mistake through its deeper understanding of cross-slot relations. In the second case, the model has made predictions for the hotel type and area even though the dialogue does not mention a hotel. This happens because the model has learned to predict the same type of slots for the attraction domain and has overfitted them during training. In contrast, Prompter ameliorates this form of over-prediction (\S \ref{Sec:fineg}). 

Our model has a weakness when it comes to dealing with slots that are unique and do not have similar slots in the source domain. In the third case, the model struggles to accurately predict the `hotel-type' and `hotel-internet' slots because they are dissimilar to all slots in the source domain.

\subsection{Why Prefix-Tuning?}

\begin{table}[]
\centering
\begin{tabular}{|c|c|c|}
\hline
           & T5DST* & \begin{tabular}[c]{@{}c@{}}T5DST* + \\ Prompt-tuning\end{tabular} \\ \hline
Attraction & 31.68 & 31.68               \\ \hline
Hotel      & 18.51 & 15.4                \\ \hline
Restaurant & 19.66 & 18.23           \\ \hline
Taxi       & 64.77 & 64.71               \\ \hline
Train      & 33.5  & 35.4               \\ \hline
\end{tabular}
\caption{Zero-shot joint-goal accuracy (\%) results on MultiWOZ 2.1 dataset comparing T5DST baseline with prompt tuning version of our approach. Both models are trained using T5-small model and semi-frozen training scheme. The results are averaged over three runs.}
\label{Tab:prompt_vs_prefix}
\end{table}

We also try implementing Prompter using soft prompt-tuning rather than prefix-tuning. Under this setting, the learned prompts are fed directly at the input layer instead of as prefixes to the attention mechanism at each layer. We compare the performance of this method with the baseline T5DST, using T5-small as the language model. We find that prompt-tuning is not even comparable to the fine-tuning baseline let alone to prefix-tuning, \textit{c.f.}~\Cref{Tab:prompt_vs_prefix}. We believe this difference is due to the fact that prompts fed in the initial layer of the transformer have a diminishing effect on the output of the decoder. This is also evident in the original prefix-tuning paper where~\citet{li-liang-2021-prefix} claim it performs better compared to prompt-tuning when it comes to generation tasks.

\section{Conclusion}
Parameter Efficient Transfer Learning methods have been frequently used for their strong robust features under a low-resource setting. However, there is no straightforward way to take advantage of these features under a zero-shot setting because they require at least some supervised data during adaptation.  The dialogue state tracking (DST) task, on the other hand, has just the right annotation for this scenario as it contains schema annotations with slot label descriptions. We propose Prompter, which uses these descriptions to enable prefix-tuning, a well-known PETL method, for use under a zero-shot domain adaptation setting. 

We show through experiments that this method improves the JGA metric for the two most common DST benchmarks. We further explain through analyses and a case study that the reason behind the Prompter's power is two-fold. (1) It has better capability to distinguish `none' valued slots within the dialogue and (2) it can digest the frequency of slots co-occurrences within the dialogue context into the prefix generation process. We believe that this study shows PETL's hidden potential for DST domain adaptation under a zero-shot setting. 

\section{Acknowledgements}
This research was supported by the SINGA scholarship from A*STAR. We would like to thank anonymous reviewers for their insightful feedback on how to improve the paper. 

\section{Limitations}
One limitation of our study is that we only evaluated our method on the T5 architecture. Further experiments on other architectures could be useful to determine the generalizability of our findings. Additionally, as in previous SOTA, our model also did not produce better results for the hotel domain, even though it did improve performance in general. We have attempted to explain why this domain is more difficult, but more research is needed to fully understand the reasons for this variability and to create methods that can improve performance across all domains.

\bibliography{anthology,custom}
\clearpage
\appendix
\section{Prefix Heatmaps}
\label{App:prefix}
\begin{figure}[h]
    \centering
    \includegraphics[width=.4\textwidth]{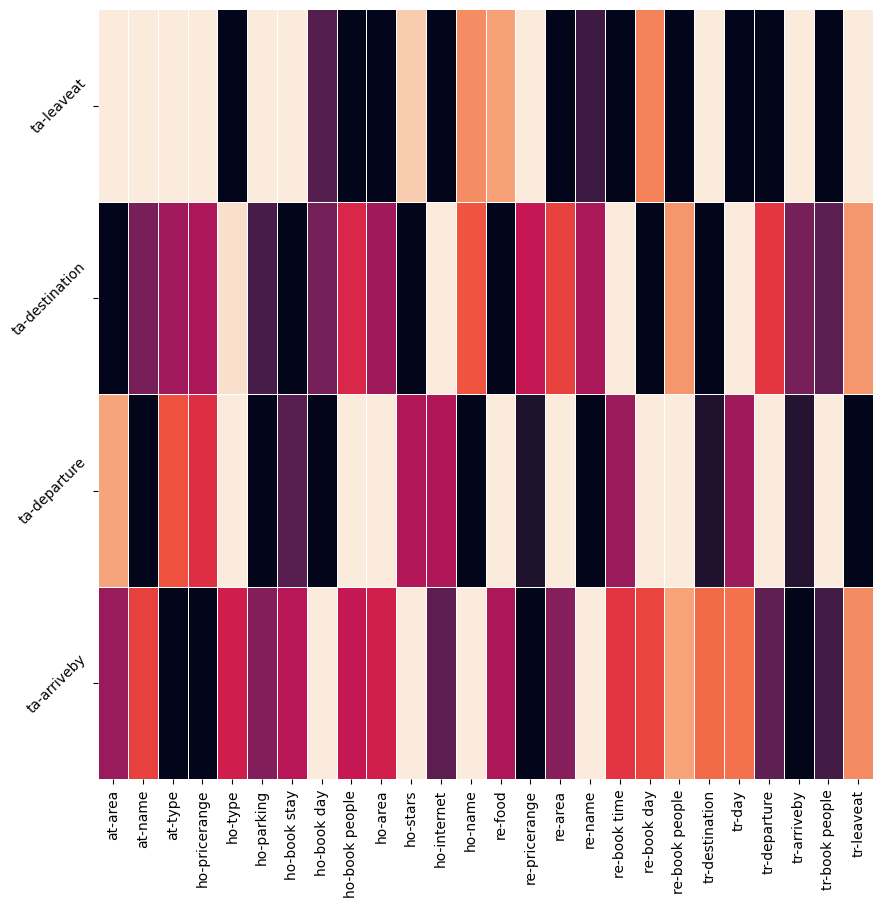}
    \caption{Heatmap for Taxi domain slots.}
    \label{fig:apptaxi}
\end{figure}

\begin{figure}[h]
    \centering
    \includegraphics[width=.4\textwidth]{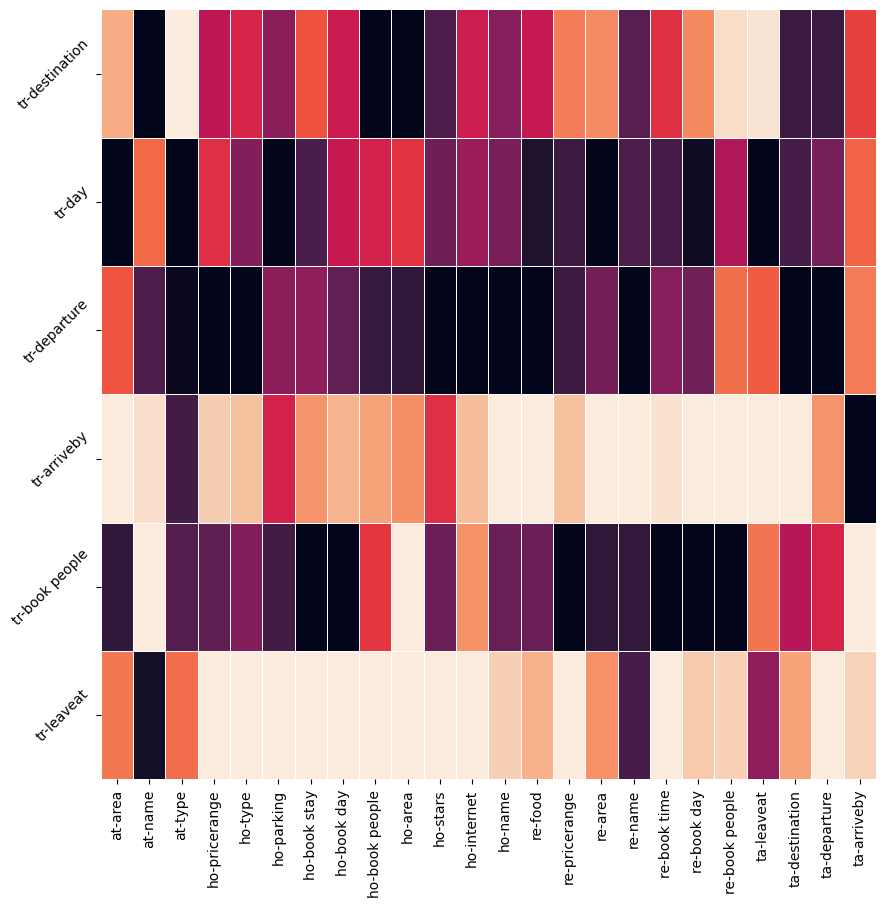}
    \caption{Heatmap for Train domain slots.}
    \label{fig:apptrain}
\end{figure}

\begin{figure}
    \centering
    \includegraphics[width=.4\textwidth]{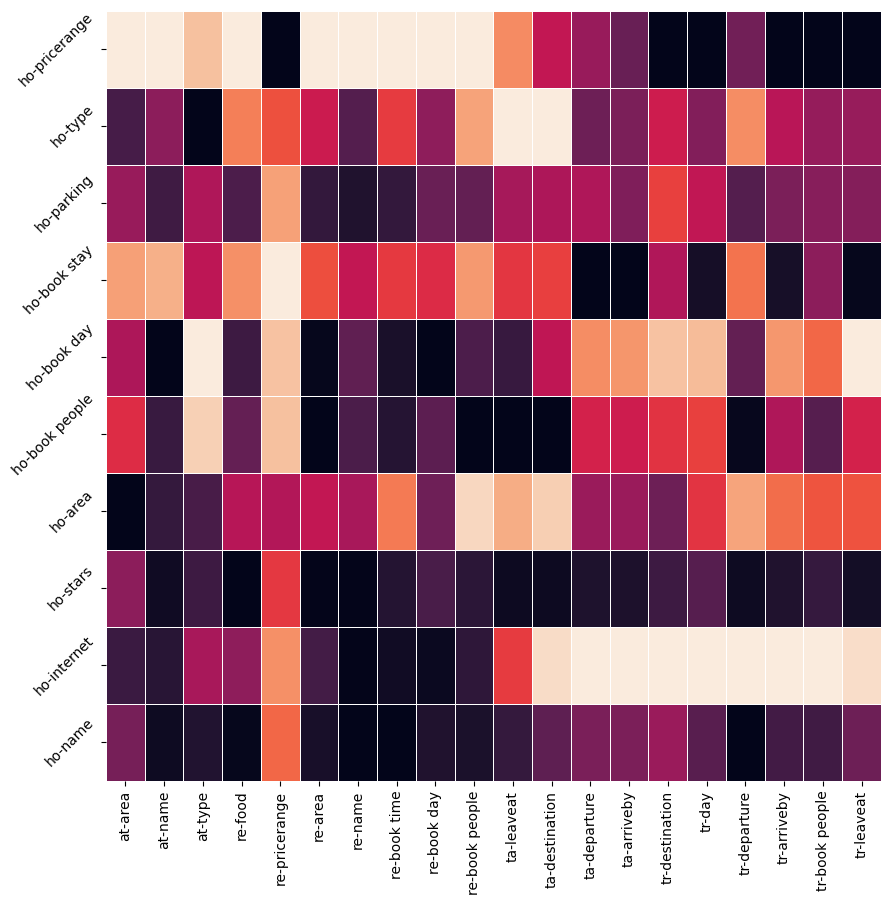}
    \caption{Heatmap for Hotel domain slots.}
    \label{fig:apphotel}
\end{figure}

\begin{figure}
    \centering
    \includegraphics[width=.4\textwidth]{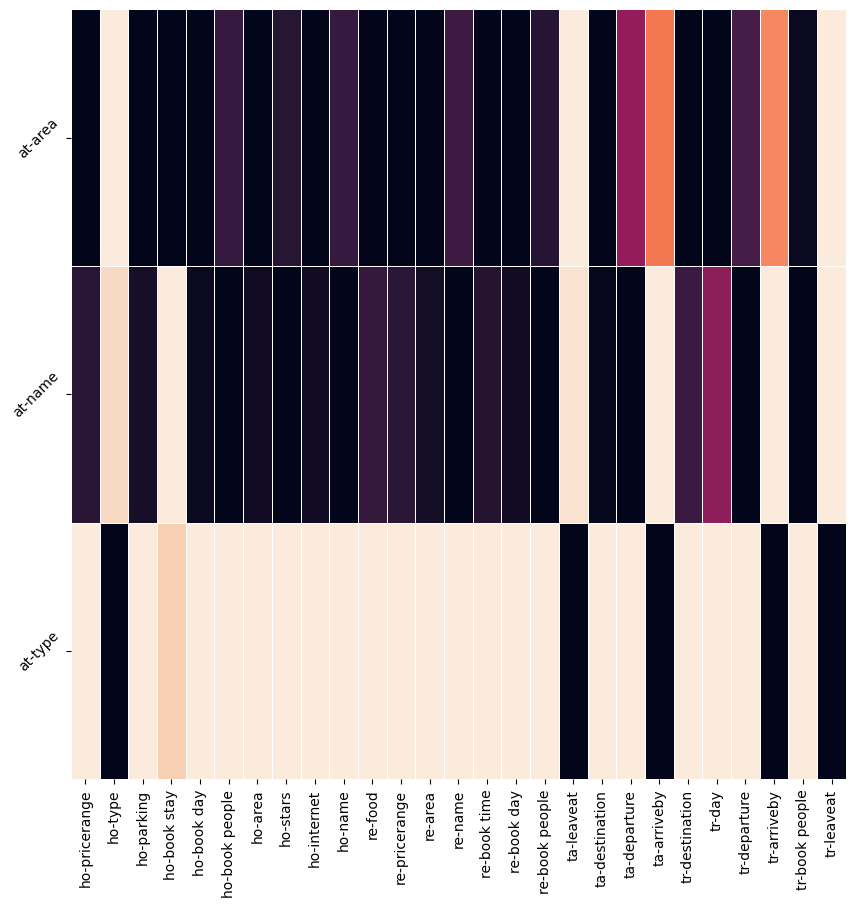}
    \caption{Heatmap for Attraction domain slots.}
    \label{fig:appattraction}
\end{figure}

\begin{figure}
    \centering
    \includegraphics[width=.4\textwidth]{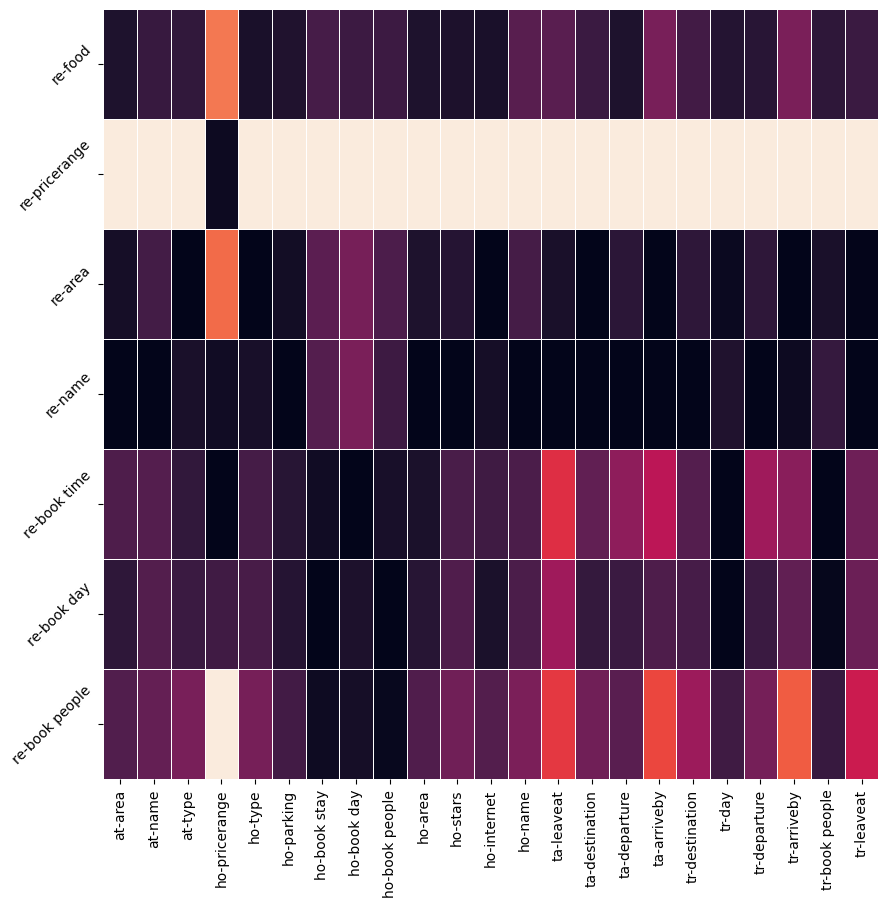}
    \caption{Heatmap for Restaurant domain slots.}
    \label{fig:apprest}
\end{figure}

Figures~\ref{fig:apptaxi} to \ref{fig:apprest} depict cosine similarity heatmaps between target and source domain slot prefixes for every domain in Multiwoz 2.1 dataset.

\section{Semi-Frozen Training}
\label{sec:semi_frozen_search}
After discovering that completely freezing the parameters of the Language Model (LM) does not lead to improved performance in zero-shot adaptation, we conducted a series of initial experiments to determine the most effective configuration. These preliminary experiments focused on the train domain of MultiWOZ 2.1. Each experiment involved training all parameters for 1,000 steps, which consistently showed benefits. We then selectively froze layers, with the specific layers varying for each row in \Cref{Tab:semi_frozen_search}. For example, in the first row, we froze all layers except the first layer of the encoder and decoder after the initial 1,000 steps. Our findings revealed that the optimal approach is to freeze all layers except the first and last layers of both the encoder and decoder after 1,000 steps.
\begin{table}[]
\centering
\small
\begin{tabular}{|l|l|}
\hline
Setting                                & JGA  \\ \hline
Unfreeze all up to 2nd layer               & 37.9 \\ \hline
Unfreeze all up to 3rd layer               & 37.7 \\ \hline
Unfreeze all up to 4th layer               & 38.8 \\ \hline
Unfreeze all up to 5th layer               & 34.5 \\ \hline
Unfreeze all up to 6th layer               & 39.1 \\ \hline
Unfreeze the first and last layers (ours)         & \textbf{39.7} \\ \hline
Unfreeze the first two and last two layers & 30.2 \\ \hline
\end{tabular}
\caption{Zero-shot joint-goal accuracy (\%) results on MultiWOZ 2.1 dataset, train domain using Prompter. Each row uses a different configuration of the semi-frozen training scheme.}
\label{Tab:semi_frozen_search}
\end{table}

\end{document}